# Semi-supervised Identification and Mapping of Surface Water Extent using Street-level Monitoring Videos

Ruo-Qian Wang[1][*], Yangmin Ding[2]

[1]*Department of Civil and Environmental Engineering, Rutgers, The State University of New Jersey, RWH 328E, 500 Bartholomew Road, Piscataway, NJ, USA 08854,*

[2]*NEC Laboratories America Inc., 4 Independence Way, Princeton, NJ 08540, USA*

**Abstract**

Urban flooding is becoming a common and devastating hazard to cause life loss and economic damage. Monitoring and understanding urban flooding in the local scale is a challenging task due to the complicated urban landscape, intricate hydraulic process, and the lack of high-quality and resolution data. The emerging smart city technology such as monitoring cameras provides an unprecedented opportunity to address the data issue. However, estimating the water accumulation on the land surface based on the monitoring footage is unreliable using the traditional segmentation technique because the boundary of the water accumulation, under the influence of varying weather, background, and illumination, is usually too fuzzy to identify, and the oblique angle and image distortion in the video monitoring data prevents georeferencing and object-based measurements. This paper presents a novel semi-supervised segmentation scheme for surface water extent recognition from the footage of an oblique monitoring camera. The semi-supervised segmentation algorithm was found suitable to determine the water boundary and the monoplotting method was successfully applied to georeference the pixels of the monitoring video for the virtual quantification of the local drainage process. The correlation and mechanism-based analysis demonstrates the value of the proposed method in advancing the understanding of local drainage hydraulics. The workflow and created methods in this study has a great potential to study other street-level and earth surface processes.

[*]Corresponding author.
 *Email addresses:* rq.wang@rutgers.edu (Ruo-Qian Wang), yding@nec-labs.com(Yangmin Ding),

## 1. Introduction

In the most cities of the United States, urban flooding is becoming more common and destructive to the society, causing loss of lives, economic damage, social disruption, and housing inequity. As a major contributor, urban flooding contributes to the general flooding costing the damage of $9 billion and 71 lives annually (The National Academies of Science, Engineering, and Medicine, 2019). Communities across the country are facing similar challenges and the increasing trend will continue with the growing number of extreme weathers and changing climate.

Compared to other types of flooding, urban flooding is known difficult to monitor and quantify, so the understanding of its drainage process, especially on the local scale, is highly limited due to data availability and measurement issues. Anthropogenic drainage systems are primarily designed to mitigate flood risks (ASCE, 1992). Despite their importance in managing flooding events, there is little information available regarding the interaction between the runoff, the paved surface, and the design of the drainage systems. A major reason of the knowledge gap is the intricate mechanism of the flood hydrodynamics in the urban environment involving numerous meteorological and hydrological factors such as the distribution of precipitation, the wide geospatial extent and variety of flooding events, the soil moisture inherited from the past rainfalls/floods, the local configuration of drainage systems, and the terrain and landscapes. So collecting critical data to understand urban flooding requires a monitoring system of wide coverage and high resolution of local details capturing the mentioned features.

The bottleneck that prevents developing and improving urban flood models and forecasting systems is the scarcity of data. Urban flooding events, especially those less dramatic ones, are poorly documented (Galloway, et al., 2018). This data scarcity problem is partly due to 1) the high cost of sensing network installation and maintenance for the wide urban area and the traditional sensor-based water level measurement, which is usually designed for deep water flows and limited to sparse fixed locations, 2) the technical issues to address in remote sensing, (e.g., satellite imaging is affected by cloud cover and complex street geometry, and its revisit interval is too long to capture the flooding event), and 3) the labor costs associated with traditional geological surveys, which are targeted to collect high water mark data after floods. These knowledge and data gaps prevent researchers from systematically examining the events, reliably identifying the driving mechanisms, and effectively developing numerical models. Consequently, decision makers cannot be informed about flood mitigation measures, flooding risks, and prevention strategies.

The recent development of smart city technology is providing an exciting new opportunity to advance the data collection and understanding of urban floods. Across the urban area, various traffic cameras are pre-existing to monitor the traffic, street activities, and coastal areas to provide a valuable data source to track urban floods (Jongman et al., 2015; Wang et al, 2018; Huang et al., 2019; Hu and Wang, 2020; Wang et al., 2020; Fan et al., 2020). Smart phones have fundamentally changed the lifestyle and business in the world and are becoming a reliable means of citizen science-based data collection, e.g., smartphone Apps and social media were used to collect critical and real-time flooding data (Wang et al., 2018). These emerging, visual data-based methods could potentially cover a large area and recognize shallow water accumulations with affordable costs and abundant details. However, analyzing such visual data is challenging. First, it is difficult to recognize the water pixels in the video or photos because the varying weather condition and daylighting lead to a changing illumination for the monitored scene so that the water ponds usually have fuzzy boundaries; the background could significantly change to prevent reliable object-based image analysis; AI and computer vision-based segmentation algorithms require dense labeling data as well as heavy computational loads to achieve an acceptable level of accuracy. In addition, these street-level data sources usually have oblique angles and distorted imagery, which pose further challenges in data processing comparing to the traditional rigorous scientific measurements. The present study is designed to develop a new semi-supervised flood water recognition scheme, which requires a low amount of labeling effort, coupling a visual data projection method called monoplotting to allow a high quality and georeferenced data processing.

Along with the smart city development, it is also worth mentioning that emerging sensing technologies have a great potential to revolute the urban flooding sensing. For example, the pre-existing optical fiber network that is originally deployed for telecommunication exhibits high sensitivities to external physical perturbations such as temperature, strain, acoustic vibration, and pressure. This new technology could be used to enable wide and high-definition sensing of urban flooding with minimal capital investment. Since this type of sensing is only available underground, a reliable method to correlate the water level underground and above the surface is required to enable such underground water level-based flooding sensing. However, the knowledge gap of the correlation between the surface water accumulation and the underground water level prevents the development of such technology. One purpose of the presented study is to fill the gap by developing the necessary knowledge to enable researchers to collect the critical data and validate the involved methods using a field study.

This paper presents a monitoring video and semi-supervised segmentation-based street-level urban flooding image analysis. A series of new methods are developed and applied to address the challenges. Section 2 introduces the work related to this study including segmentation, monoplotting, and their application to urban floods. Section 3 describes the photogrammetry methods and the details of the field experiment. The validation of the methods and analysis results are presented in Section 4. The study is summarized at last.

## 2. Related Work

This study developed a novel framework of street-level flood recognition and mapping using machine learning and photogrammetry techniques. In this section, we first review the related machine learning methods with a focus on semi-supervised learning and the progress of monoplotting. Then, we will review the application of these methods to flood monitoring and warning.

### 2.1 Machine Learning Based Object-based Image Analysis

Object-based Image Analysis (OBIA) is a technology to classify image pixels into segments and measure physical quantities based on the relative position between the classified segments and the background. OBIA has been used to analyze aerial photography for vegetation and urban landscapes, natural disaster damage analysis, and risk management (Blaschke, 2010; Garcia et al., 2018; Lee and Yang 2018; Bandini et al., 2017, Van der Sande et al., 2003). Classic manual image segmentation methods (Alvarez et al., 2012; Barrow and Tenenbaum, 1981; Grady, 2006; Kass et al., 1988; Roerdink and Meijster, 2000) are often sensitive to subjective inputs and image noise, while the recent deep learning-based algorithms allow more robust extraction of visual features for segmentation analysis, e.g., the supervised learning algorithms of fully convolutional network (FCN) (Long et al., 2015), the convolutional neural network (CNN) architectures (Krizhevsky et al., 2012), VGG16 (Simonyan and Zisserman, 2014), ResNet (He et al., 2016), and the compound methods by Hariharan et al. (2014), Arnab and Miksik (2017), Bai and Urtansun (2017), and Liu et al. (2017), which combined multiple methods for segmentation. But these supervised methods still require heavy labeling burden, and they are difficult to scale up to other applications, especially in the cases where experience and deep expertise are required.

The image segmentation has been applied to analyze remote sensing data in urban studies. A detailed review focusing on land cover can be found in Ma, et al. (2017) and Hossain and Chen (2019). However, these classical segmentation algorithms, such as region growing (Jasiewicz et al., 2016), region merging (Vasuki et al., 2017), and clustering (Chen et al., 2018), are insufficient and unreliable in analyzing water data that contains heterogeneous textures and uneven/low illumination. Deep learning-based image segmentation can, to an extent, address the technical issues but they show limitations in recognizing fuzzy water accumulation boundaries under the natural lighting conditions and suboptimal background contracts, making segmentation from a single scene extremely challenging. The labeling burden in such tasks is also daunting as the dramatically changing environment requires dense label data, which could be generated through a tedious, costly, and inaccurate process. The background change such as the reference object motion adds extra errors to the object-based measurements. For example, Jafari et al. (2021) applied segmentation to estimate flood levels in downtown Houston, Texas during Hurricane Harvey and found the weather and background motion introduced nontrivial errors to the analysis.

Semi-supervised methods, which could alleviate the burden of manual labeling, are gaining a rising attention recently. In this approach, a small set of images are fully annotated but the majority of the image dataset are unavailable. The typical semi-supervised methods include creating surrogate classes (Pickard, 2002), adding entropy regularization (Kull, 2005), using

Generative Adversarial Networks (GANs) (Svenningsen et al. 2015), or averaging ensemble networks (Gimmi et al. 2016; Perone and Cohen-Adad, 2018). Among these methods, self-training is considered an emerging and future machine learning strategy, in which a network is created using the labeled data to train a preliminary network, which is used to predict the labels of the unlabeled data to produce the so-called pseudo labels. Then, the network is retrained using the augmented training set combining the manual labels and the pseudo labels for improved accuracy (Bai, 2017; Pathak et al, 2015; Rajchl et al., 2017). Although this approach can leverage unlabeled images, mistakes made in the early training process could propagate back to the network and be amplified during training (Chapelle, et al., 2006; Zhu and Goldberg, 2009). Despite several techniques were proposed to overcome this issue, self-supervised learning is still challenging to achieve a high level of accuracy and the training process can be difficult to control to gain the convergence.

Monitoring footage contains visual data of strong temporal coherency and continuous deformations, so more features can be leveraged to enhance the accuracy of segmentation, e.g., the Deep Feature Flow (Zhu et al., 2018), the post-segmentation refinement (Chen et al., 2018a; Lin et al., 2016; Zhu et al., 2018), Feature Space Optimization (Kundu et al., 2016), and Gated Recurrent Flow Propagation (Nilsson and Sminchisescu, 2016). However, these methods need dense labeling data (e.g., pixel-wise labeled videos), which could be unaffordable for practical applications. The semi-supervised segmentation method that has less labeling burden has much less been applied to analyze video segmentation and the present study is aimed at filling the gap for a specific application in urban flooding.

**2.2 Monoplotting**

The street-level imagery obtained from smartphones, UAV (Unmanned Aerial Vehicle), and urban and coastal monitoring cameras, tend to provide photos and videos at oblique angles. To retrieve valuable data from such emerging data sources, we need to break through the barrier posed by oblique imagery data. Monoplotting is a potential solution that allows the users to obtain image registration to the 3-D Digital Elevation Model (DEM). Monoplotting was named by Makarovic (1973), who used this method to transform photographs into georeferenced data (Bayr, 2021). This method has several applications and advantages in analyzing geoscience problems: First, the image registration enables the geo-locating of the captured objects and process in the DEM (or a map) for image content quantification. Second, monoplotting requires only a single image and a DEM to obtain 3-D observation of the scene to avoid the data requirement of stereo images to reconstruct 3D information. Monoplotting has been used in extracting long-term earth surface changes using repeat photography (Pichard, 2002; Kull, 2005; Svenningsen et al., 2015; Bayr and Puschmann, 2019), obtaining 3-D view of the scene with the ground-based photographs, and analyzing historical photographs from the time before aerial photography appeared (Gimmi et al., 2016). Recently, this method shows usefulness in the analysis of crowdsourcing images to study emergent and rapid processes such as flooding (Golparvar and Wang, 2020; Triglav-Čekada and Radovan, 2013) and the potential to use the oblique data to communicate with stakeholders taking the advantage of the similarity with daily-life perception and experience (Triglav-Čekada et al., 2011). Low cost sensing technology also benefited from this powerful georeferencing method, e.g., in the monitoring of slow earth surface dynamics such as glacier movement and vegetation changes (Triglav-Čekada et al., 2011; Triglav-Čekada et al., 2014; Wiesmann et al., 2012). Dynamic information obtained in the video could be used to reconstruct 3- D deformation and movement such as in the dam breaking event for forensic analyses (Travelletti et al., 2012; Yuan et al., 2021) and multitemporal landslide monitoring (Makarovic, 1983).

Recognizing the great value of monoplotting, handy software and codes have been created for the geoscience community, including the OP-XFORM project (Aschenwald et al, 2001), the JUKE method (Corripio, 2004), Georeferencing oblique terrestrial photography (Mitishita et al., 2004), the 3D Monoplotter (Fluehler et al., 2005), and the DiMoTeP (Marco et al., 2018). Recently, the WSL Monoplotting Tool (WSL-MPT) developed by the Swiss Federal Research Institute (WSL) is gaining popularity and has been applied to the quantitative analysis of natural hazards (Triglav-Čekada et al., 2011; Scapozza et al., 2014), glacial processes (Stockdale, 2015), and land cover changes (Gabellieri and Watkins, 2019; Stockdale et al., 2019; McCaffrey and Hopkinson, 2017). A similar tool called Pic2map, leveraging the convenience of the Geographic Information System (GIS) software QGIS, shows a strong rising trend (McCaffrey and Hopkinson, 2020). The present study used Pic2map for the following analysis.

**2.3 Computer Vision-based Flood Data Mining**

Computer vision and machine learning have been applied to analyze street-level flood data. For example, citizen science-based disaster surveillance has been an emerging and cost-effective data collection method to provide wide geospatial coverage with high resolution (Savage, 2013). UAVs are becoming increasingly popular among citizens, organizations, and researchers to make field observations of flood events (Golparvar and Wang, 2020), which provides outstanding resolution, wide-coverage, and flexible monitoring time. CCTV cameras provide continuous, high resolution, and consistent monitoring data but the location of such data is fixed for narrow coverage and the footage quality is sometimes poor in adversarial weather condition.

The early machine learning's applications to visual data focused on the image level recognition of social media data and georeferencing the data with the related text messages (Wang et al., 2018 and Wang, 2018). The method was extended to classify the images into four categories, i.e., first-hand witness, preparation, recovery, and public reports, to create a real-time flood warning and emergency response information system (Wang et al., 2020). The introduction of AI enables the automatic extraction of new information that the traditional manual data analysis cannot compete in terms of processing speed, data volume, and quality.

Segmentation has been used to perform a more accurate measurement of flood extents and depths. This process is traditionally accomplished for satellite imaging (Nemni et al., 2020) through image vectorization, which requires drawing of punctual, linear, and polygonal elements at the photograph and exported as layers in GIS software and overlaid with cartographic elements (Gabellieri and Watkins, 2019). Since the traditional methods such as thresholding and region growing are shown less effective than deep learning methods (Arshad et al., 2019), we, therefore, focus on reviewing the latter, which could largely improve the time and labor input in such tasks. Chaudhary et al. (2019) applied CNN to automatedly classify the water submergence segmentations into ten levels based on multiple objects in social media images to estimate flood depth. Javid et al. (2020) applied segmentation to recognize the water area in monitoring videos and used reference objects to estimate the water level for the flood of Hurricane Harvey. Moy de Vitry et al. (2019) presented a flood level trend monitoring approach through detecting floodwater in surveillance footage of a camera system with a deep CNN. Additionally, Chang et al. (2019) reviewed various machine learning methods applied in flood forecast modeling and reported that machine learning methods are the key in developing early warning systems for urban flood hazards. These studies demonstrated the high portability of the CNN in image detection to be applied in many fields of flood monitoring, modeling, and forecasting. However, they found that these methods could be significantly impacted by the distortion and mirror reflection of the image (Song and Tuo, 2021), and they could be difficult to scale due to the high cost of infrastructure and maintenance constrains to derive precise flood depth data (Moy de Vitry et al., 2019). In addition, the data quality could be compromised due to the difficulty in georeferencing (Golparvar and Wang, 2020).

On the other hand, the value of monoplotting in georeferencing and correcting oblique angle and image distortion has not been fully recognized in the flood analysis community. For example, to estimate the level of flooding presented in the monitoring video for early warning purpose, Moy de Vitry et al. (2019) developed the Static Observer Flooding Index (SOFI), which counts water pixels and their ratio in the full image, to design the trigger to send the early warning signal. As the observation angle is oblique and the pixels of the far field cover larger areas than the near field, this method could be biased to underestimate the flooding extent in the far field. Recognizing the georeferencing problem, Golparvar and Wang (2020) collected social media photos during the high tide flood that occurred in Newport Beach, California in July 2020 and used monoplotting to georeferenc the flood extent boundary for correcting the oblique angle bias. A more extended study by Yuan et al. (2021) analyzed a dam break incident in Michigan in 2020. The collected smartphone data was used to generate a dataset of 3D georeferenced dam deformation and clasping process.

## 3. Methods and Experiments

### 3.1 Semi-supervised Image Classification

Recognizing the boundary of water accumulation on road surface is a challenging task. The mirror reflection, oblique observation angle, and image distortion make the boundary of water accumulation fuzzy. The traditional supervised learning methods could, to some extent, address the problem but it requires detailed labels for some, if not all, the frames of the video, which is usually time-consuming, expensive, and tedious. In addition, manual labeling, which depends on the experience and judgement of the label preparer, could be inaccurate in recognizing the fuzzy boundaries.

We proposed a new semi-supervised segmentation algorithm, which can efficiently segment an image into different categories with minimal labeling inputs and effectively recognize the water accumulation boundaries without the impact of labeling accuracy. Specifically, the video is first cropped to the interested area with a matrix of $X_{M \times N \times S \times 3}$, where $S$ frames have $M$ rows, $N$ columns, and $3$ bands of pixels. An unsupervised segmentation method, i.e., Felzenszwalb algorithm (Felzenszwalb et al., 2009), was applied to cluster the pixels of a representative frame $X_{M \times N \times 3}$ into $F$ ($\ll M \times N$) segments ($\hat{X}_{T \times 3}$) to generate a reduced-order image. A sparse grid with $L$ ($\ll F$) points was developed covering the representative frame, $\hat{X}_{T \times 3}$. So the $L$ points can be seen as a far smaller subset, $\hat{X}_{L \times 3}$, of $\hat{X}_{F \times 3}$. In fully supervised learning algorithms, users must label $M \times N \times S$ pixels, while in the proposed supervised algorithm, the original image is simplified to $R$ segments and only part of the $L$ points need to be labeled. Therefore, much labeling effort is saved in this new semi-supervised method.

Among the $L$ grid points, we only labeled $L_1$ "permanent" dry points (the purple dots in Fig. 1) and $L_2$ "permanent" wet point (the red dots in Fig. 1), where $L_1+L_2<L$. So $L_0$ ($=L-L_1-L_2$) points, which are difficult to classify using human judgement or changing over time, are left to be determined. The $L_1$ and $L_2$ labeled points ($\hat{X}_{L_1}$ and $\hat{X}_{L_2}$) were then used to train a supervised classification model, i.e., a decision tree model in the present project, to classify the segments containing them into the dry and wet groups. The supervised classification model is next used to predict the $P$ ($=F-L_1-L_2$) segments that are not labeled. In this approach, the segmentation only relies on the input of the users that have the highest confidence ($\hat{X}_{L_1}$ and $\hat{X}_{L_2}$), so the subjectivity and uncertainty of the labeling input is minimized. The methodology is detailed in Fig 1. In general, the labeling task is reduced from $M \times N \times S$ to $L_1+L_2$. As an example, the field experiment described below had a video of $M \times N \times S = 1072 \times 1920 \times 145 \approx 3 \times 10^8$ pixels to label/predict, and they were reduced to the number of $L_1+L_2=16+92=108$ points to label.

## 3.2 Monoplotting

To obtain a wide observation range, monitoring cameras usually employ a wide-angle lens, which distorts the image and poses challenges in georeferencing. We adopted the monoplotting technique to address this issue. Ground Control Points (GCPs) were identified in the image and the DEM. Once the GCPs are determined, an optimization scheme is employed to find the location and parameters of the camera to match the two sets of GCPs from the image and DEM in the projection plane to allow the determination of the pixel coordinates in the 3D space. This involves three steps: First, the GCPs captured in the DEM (called DEM GCPs) are projected to a plane assuming a camera is positioned in a location ($x, y, z$) using the equation below,

$$\begin{bmatrix} u \\ v \\ w \end{bmatrix} = K_{3\times3} R_{3\times3} \begin{bmatrix} X + T_x \\ Y + T_y \\ Z + T_z \end{bmatrix}_{3\times1} \tag{1}$$

$$R_{3\times3} \begin{bmatrix} X + T_x \\ Y + T_y \\ Z + T_z \end{bmatrix}_{3\times1} = \begin{bmatrix} R_{11} & R_{12} & R_{13} \\ R_{21} & R_{22} & R_{23} \\ R_{31} & R_{32} & R_{33} \end{bmatrix} \begin{bmatrix} X + T_x \\ Y + T_y \\ Z + T_z \end{bmatrix} = \begin{bmatrix} R_{11}(X + T_x) + R_{12}(Y + T_y) + R_{13}(Z + T_z) \\ R_{21}(X + T_x) + R_{22}(Y + T_y) + R_{23}(Z + T_z) \\ R_{31}(X + T_x) + R_{32}(Y + T_y) + R_{33}(Z + T_z) \end{bmatrix} \tag{2}$$

where $u$ and $v$ are the horizontal and vertical coordinate of projected points in the image plane, $K$ is the intrinsic camera matrix in which there are at least 4 unknown parameters such as focal length and image center coordinates. $R$ is the rotation matrix and $T$'s components are for the translation vector. Second, the projected GCPs are compared with the GCPs identified in the image (called Image GCPs). Finally, the camera parameters and location are iteratively adjusted to reproject the DEM GCPs until the best comparison is achieved. The real-world coordinates obtained through monoplotting are then used to project the recognized water accumulation distribution to the DEM or a map. This georeferenced data will allow the measurement of the water areas in the real-world scale. The monoplotting operation was conducted with Pic2map (McCaffrey and Hopkinson, 2020).

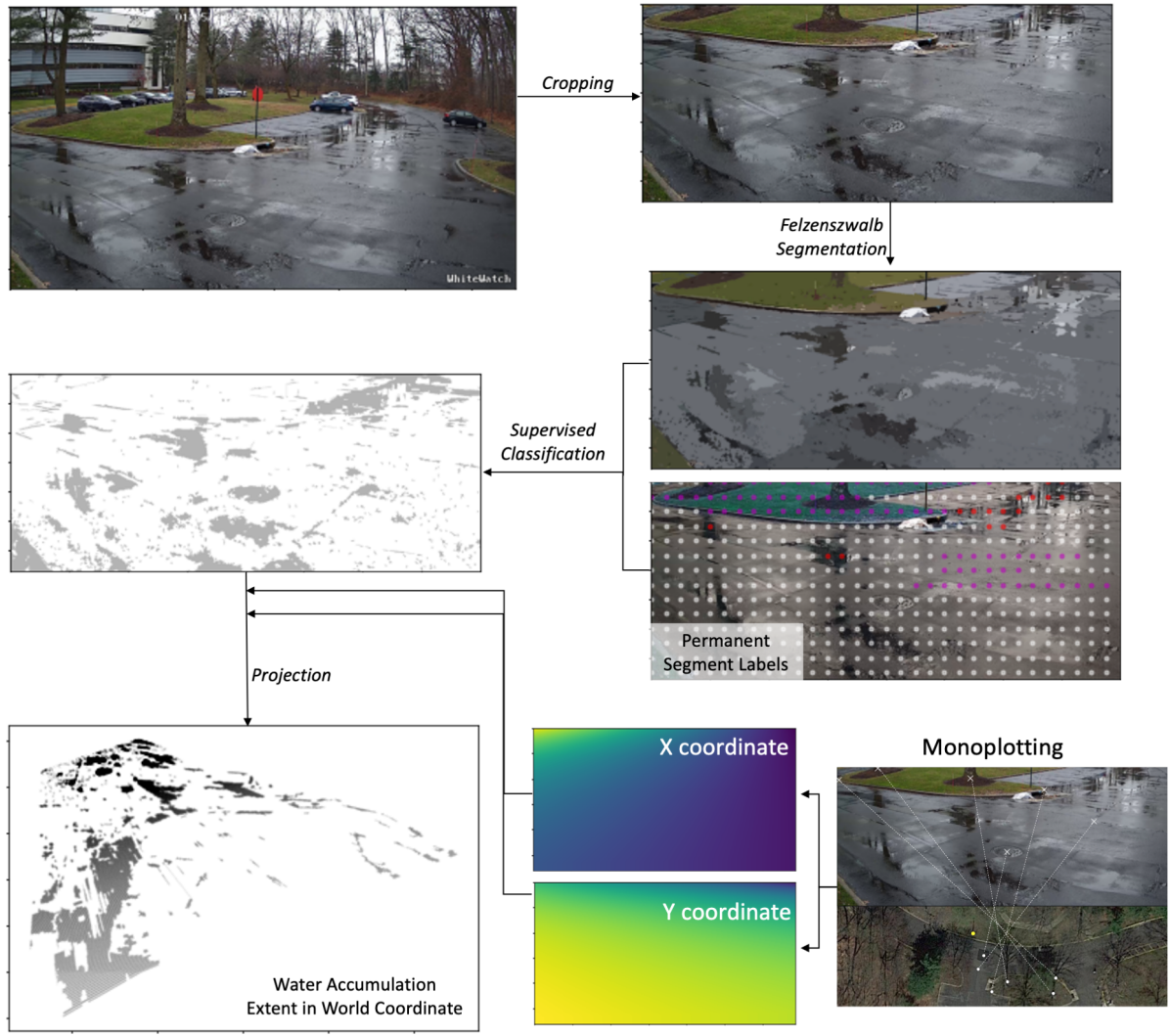

Figure 1 Overview of the data analysis method

**3.3 Field Experiment**

A field experiment was conducted in January of 2022 at a parking lot in Princeton, New Jersey. A solar-panel powered video camera (Reolink Go 4G) was installed on a light pole. This camera transmited data through the wireless 4G network and recorded the field every 30 seconds. The camera was zoomed to a drainage well to collect the runoff of the area in the view of the camera. The water level inside the well was connected to an outlet during the experiment. The water level inside the well was monitored by an ultrasonic liquid level transmitter with an accuracy of 0.5% of the measurement value. The relative position of the camera to the drainage well is shown in Fig. 2.

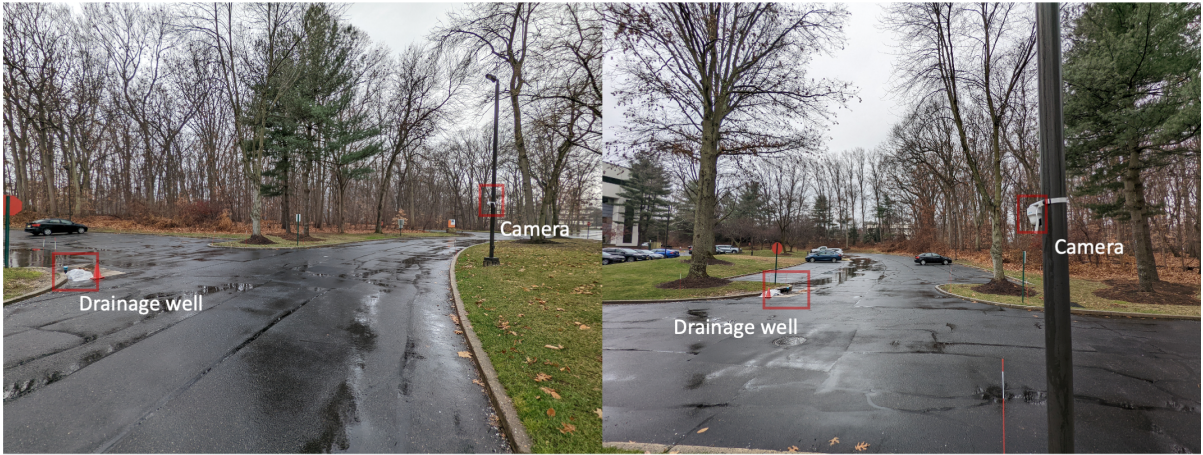

Figure 2 The setup of the field experiment.

Sample frames from the captured video are shown in Fig. 3. These images show that the water accumulation surface had strong mirror reflection of the surrounding environment such as treets and cars. The image illumination contract varied over time depending on the cloud cover and weather change. The background objects such as human and cars had movement and the camera was sometimes blocked by the ongoing traffic. These factors contribute to the uncertainty of the data analysis. The DEM data in 10-meter resolution was collected through the database of New Jersey Department of Environmental Protection Open Data Webpage (https://www.state.nj.us/dep/gis/wmalattice.html) to support the monoplotting operation. Note that monoplotting was originally designed to work with DEM data, but the DEM data is flat for the monitored parking lot and the geospatial scale is too low to capture any features in DEM. We used a satellite image from Google Map overlaying on the DEM data so that the corresponding

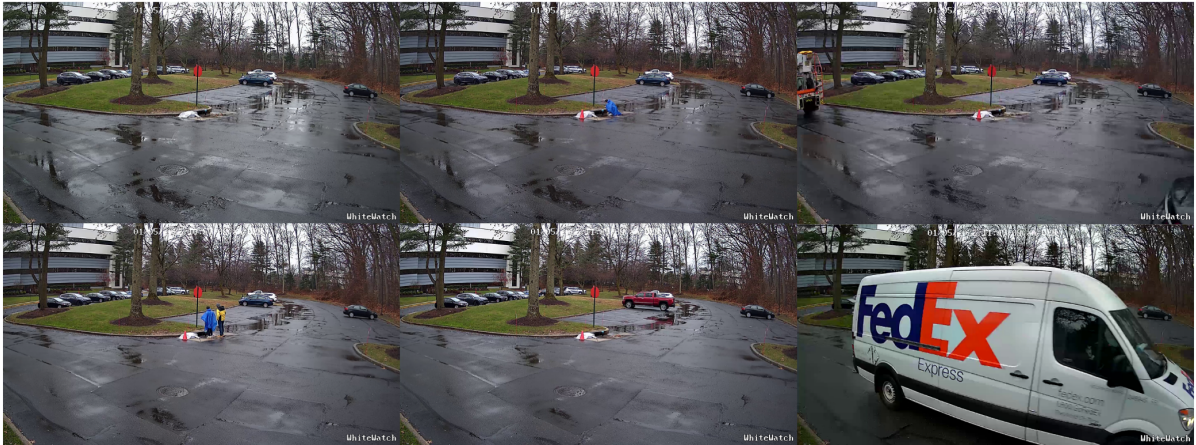

Figure 3 Sample frames of the captured video, in which objects were observed changing and moving in the video with varying illumination and camera blocking.

features could be captured.

4. **Results**

Sample segmentation results along with their original images are shown in Fig. 4 for comparison. The grey areas identified as water pixels have a good coverage of the water accumulation areas in the original images. Furthermore, the proposed segmentation method captures the detailed, small water surfaces that human labeling could easily ignore. Miss-classification was observed at the edge of the curb where a dark lighting condition was present. Overall, the segmentation method has a satisfactory performance.

The monoplotting method was performed following the illustration in Fig. 5. GCPs were identified at the features in the original image and the corresponding points were determined in a map. Their correspondence was shown in Fig. 5 and we found the camera position and pose could be reliably obtained as the yellow point. Using the obtained geographic coordinates (Fig 2.), we reproject the identified water accumulation to the real-world coordinate system and a sample flood distribution and the obtained pixel-level geographic coordinates are shown in Fig. 2.

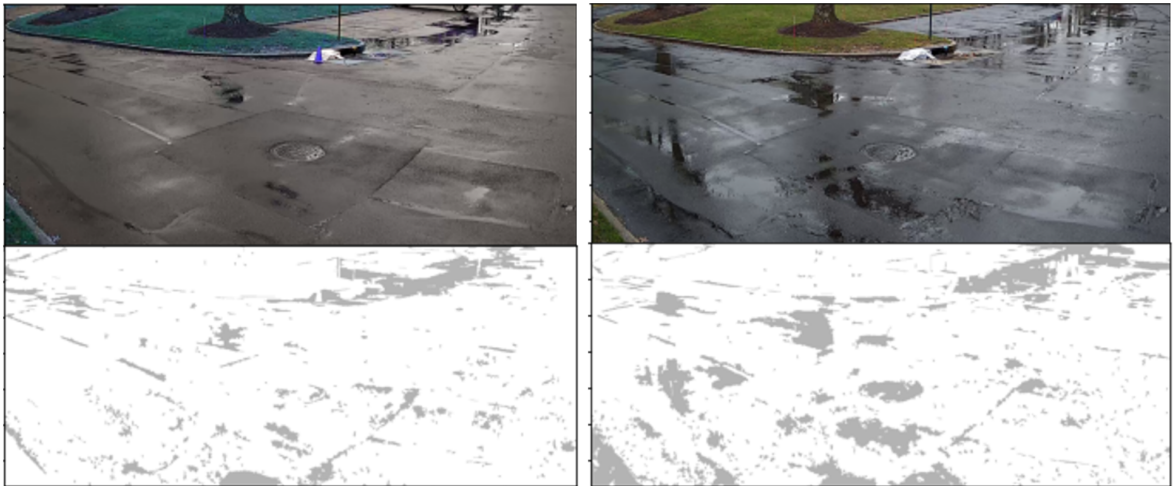

Figure 4 Sample classification results: the upper panels are the original image and the grey masks in the lower panels are the segmentation results

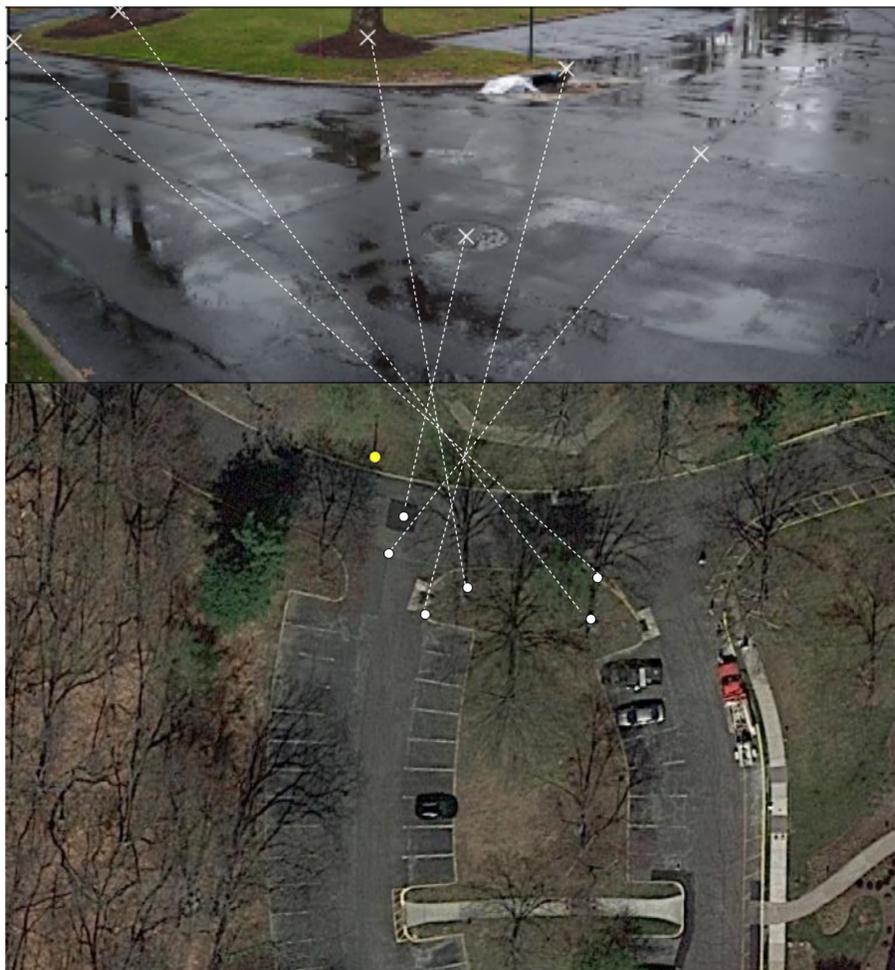

Figure 5 The identified GCPs in the original image and the map. The white dash lines denote the corresponding relationship.

From the obtained segmentation results, we can obtain two types of water accumulation distribution estimates. The first is similar to the SOFI (Moy de Vitry et al., 2019), i.e., the ratio of the water pixel number to the total pixel number. As discussed, this index could bias the estimate of flooding situation because every pixel is treated equally. So we developed the second index, which calculates the ratio of the water area to the total area in the projected 3D coordinate. The two indices for the video are shown in Fig 6. Due to the heavy uncertainty in the data collection and processing. These two data have strong fluctuation over time. The 5-min moving average scheme was then applied to remove the strong fluctuation. It is clear that the pixel-base SOFI is lower than the projected SOFI. This is expected because each of the far field pixels covers a larger area than the near field. As the far field has more water accumulation, the pixel-based SOFI underestimates the ratio of water coverage.

In comparison, the water level data collected inside the drainage well is shown in Fig. 6. The water depth value was converted from the measurement of the distance from the transmitter to the water surface with an interval of 5 mins. The water depth was found to reach the peak after a peak of the water accumulation ratio. This could be attributed to the fact that the precipitation peak leads to the peak of surface water accumulation. The water on the road surface took about 10 mins to flow to the drainage well and thus the water level in the well has a response 10 minutes lagging the road surface. This observed process is sensational and indirectly proves that the data process was reliable.

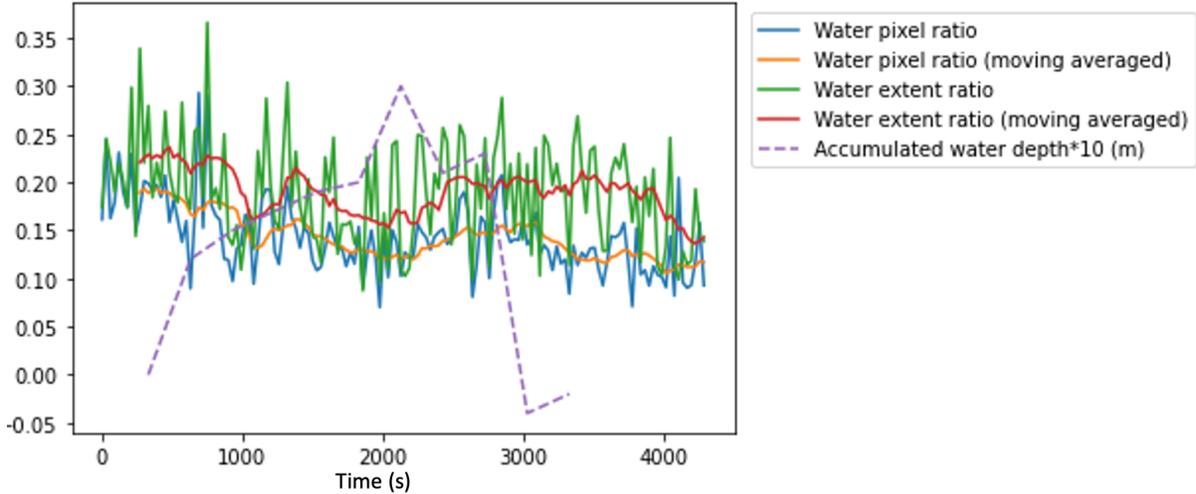

Figure 6 The water extent time series: the pixel-based and projected SOFI over time, which is compared with the water level inside the drainage well.

From the time series, we can establish a correlation relationship between the water depth in the drainage well and the water accumulation extent on the road surface. In Fig. 7, the rainfall event is divided into two phases, i.e., the rising and falling phases, based on the water level change in the well. The trend shows that at the beginning of the rainfall, the flood extent decreases when water level increases in the drainage well, whereas after the peak of the water level in the well, the flood extent starts increased and then remained the same level when water level was falling in the well. This indicates two different drainage processes in the local storm water runoff change. In the rising phase, the water accumulation was generated by direct rainfall, which flowed into the drainage well so that water level in the well increased and surface water extent decreased. In the falling phase, the surface water extent was mainly formed by collecting the surrounding storm water. In this phase, the water was mainly staying on the road surface without strong runoff to the well and the water in the well exited from the outlet in the well to cause the water level's declination in the well. The study shows that using the advanced machine learning method and monoplotting allows a detailed study in the local drainage and runoff process. The spatial and temporal resolution obtained in this study cannot be reached using the traditional methods.

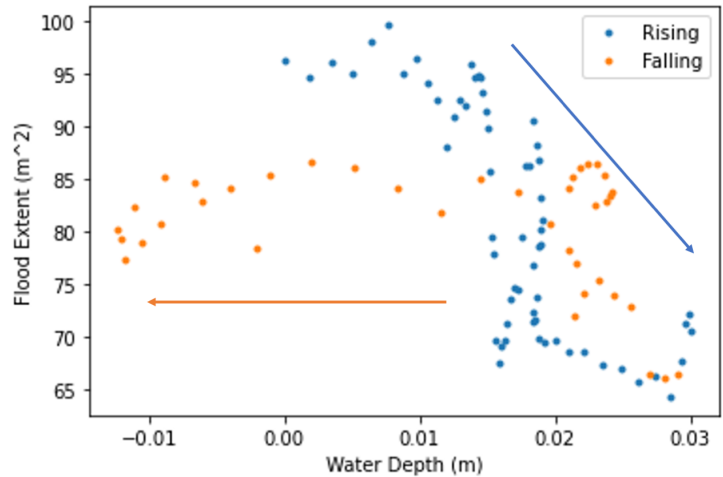

Figure 7 The correlation between the surface water extent and the drainage well water level in two phases.

## 5. Discussions

The semi-supervised segmentation method shows a good performance. It captures the water extent on the road surface with environmental uncertainties and changing background. Although the noise due to mirror reflection and illumination has been mitigated, we still observe these factors impact the final result to a non-trivial extent. The moving average of the data helps further removing such noise, but may risk losing physical information in the process. The methods involved in this study still have a potential to improve such as using more advanced unsupervised clustering schemes and supervised classification model. In addition, the resolution and level of labeling should be tested to find the best combination for accuracy.

The proposed semi-supervised segmentation method is suitable for the task to identify uncertain boundaries based on the image texture. Since only a limited places on the image are required to be labeled, this method significantly reduces the density of labeling inputs. Leveraging on the temporal coherency of videos, especially the fixed monitoring footage, the saving of labeling costs could be further amplified.

The monoplotting shows a good value to georeference the data and correct the image distortion in this study. However, identifying the GCPs needs extensive experience to reach a practical level and the matching process that depends on an optimization scheme could be difficult to converge. Iterative updates of GCPs and camera position Apriori to guide the convergence process are required. There is a hope that a full or semi-automatic monoplotting scheme could be developed to help the users to perform this task. An effort has been made in Golparvar and Wang (2021), but a large scale application and validation has not been done.

## 6. Conclusion

This study developed a novel semi-supervised segmentation scheme to recognize the boundary of surface water accumulation. The method involves an unsupervised segmentation step to reduce the dimension of clustering, a manual input to label the "permanent" segments, and a step of supervised classification. This scheme shows a satisfactory performance and saves a significant amount of labor input. This study demonstrates that a semi-supervised segmentation is suitable for fuzzy boundary identification with changing background and image quality. The monoplotting applied in this study shows a good value to reproject the image information to the 3-D real-world coordinate, which enables the quantification of the image information using photogrammetry methods.

The combination of the segmentation and monoplotting methods was shown informative to study the detailed local drainage process and a sample time series and a correlation relationship were obtained for application. This study shows that the detailed local hydraulic analysis could be possible with the presented unprecedented resolution and details. Future studies are called to improve the understanding of the runoff and drainage processes to support the technology development of smart cities.

**Acknowledgements**

The first author acknowledges the funding support by the US Department of Transportation through the Center of Advanced Infrastructure Technology at Rutgers University. The authors thank the kind support of Dr. Ting Wang from the NEC Labs in this work.